\documentclass[conference]{IEEEtran}
\IEEEoverridecommandlockouts

\usepackage{booktabs}
\usepackage{fancyhdr}
\usepackage{algorithm}
\usepackage[usenames, dvipsnames]{xcolor}

\usepackage{colortbl}
\usepackage{caption}
\usepackage{graphicx}
\usepackage{array}
\usepackage{adjustbox}
\usepackage{hyperref,graphicx,color,float}
\usepackage{url}
\definecolor{myblue}{RGB}{10, 150, 200}

\usepackage{cite}
\usepackage{multirow}
\usepackage{amsmath,amssymb,amsfonts}

\usepackage{graphicx}
\usepackage{textcomp}
\usepackage{comment}
\definecolor{highlightColor}{HTML}{E6FFE6}

\usepackage{algpseudocode}
\usepackage{amsmath,amssymb,amsfonts}

\usepackage{xcolor}
\def\BibTeX{{\rm B\kern-.05em{\sc i\kern-.025em b}\kern-.08em
    T\kern-.1667em\lower.7ex\hbox{E}\kern-.125emX}}

\usepackage[margin=1in]{geometry}

\algrenewcommand\alglinenumber[1]{\scriptsize #1}

\usepackage{amsmath}
\usepackage{microtype}
\usepackage{array}
\usepackage{booktabs}

\usepackage{xcolor}
\def\BibTeX{{\rm B\kern-.05em{\sc i\kern-.025em b}\kern-.08em
    T\kern-.1667em\lower.7ex\hbox{E}\kern-.125emX}}
\begin{document}

\pagestyle{plain} % Apply plain style to other pages

\title{A Clustering-Based Framework for Identifying Suspicious Trading Patterns in Capital Market\\

}

\author{\IEEEauthorblockN{Asif Zaman }
\IEEEauthorblockA{\textit{Department of Computer Science \& Engineering} \\
\textit{American International University-Bangladesh}\\
Dhaka,Bangladesh \\
25-93713-1@student.aiub.edu}
\and
\IEEEauthorblockN{Romona Magdalene Sarkar}
\IEEEauthorblockA{\textit{Department of Computer Science \& Engineering} \\
\textit{American International University-Bangladesh}\\
Dhaka,Bangladesh \\
24-93338-1@student.aiub.edu}
\and

\IEEEauthorblockN{Sabiha Khair Ohi}
\IEEEauthorblockA{\textit{Department of Computer Science \& Engineering} \\
\textit{American International University-Bangladesh}\\
Dhaka,Bangladesh \\
25-93714-1@student.aiub.edu}
\and
\IEEEauthorblockN{Iftekharul Mobin\textsuperscript}
\IEEEauthorblockA{\textit{Department of Computer Science \& Engineering} \\
\textit{American International University-Bangladesh} \\
Dhaka, Bangladesh \\
\textsuperscript{}iftekhar.mobin@aiub.edu}
}

\maketitle
\thispagestyle{plain} % Apply the footer only on the first page
\begin{abstract}
Market manipulation is the dubious practice of manipulating stock prices in order to make a quick profit, which truly degrades confidence on trading platforms. We implemented an unsupervised fraud-detection toolkit that begins with K-Means++ clustering to address this issue. A dataset of roughly one million financial transactions from 2012 to 2024 is used. In order to identify fraudulent trades and categorize them using market practice heuristic thresholds, the study suggests a clustering-based pipeline. The method  highlights 2.02\% of trades as suspicious where 51.10\% clearly indicate spoofing, 0.10\% indicate pump and dump, 0.55\% indicate insider trading, 1.43\% indicate a fake breakout, and 46.83\% are unclassified. Despite the lack of ground truth, the model's performance is confirmed by a Silhouette Score of 0.561.
\end{abstract}

\begin{IEEEkeywords}
Manipulation, Stock Market, Detection, Price, Visualization.
\end{IEEEkeywords}

\section{Introduction}
Every nation's infrastructure is extremely dependent on the state of its economy.  An essential component of the nation's economic growth is the stock market. As new products and trends shape the market, stock markets are evolving.
% \cite{b23}
The idea of a stock market is slowly gaining momentum in Bangladesh.  Despite improvements in financial literacy, DSE participation is still low.  Growth is still low because of investor mistrust, anticipated dangers, and previous market fraud incidents\cite{b4}.
% The relationship between Bangladesh's stock market and economic expansion has been studied by economists. 
 Due to its isolation from international markets the DSE has limited growth especially in times of crisis like COVID-19 \cite{b5}. 
 The Dhaka Stock Exchange shows chronic volatility clustering due to  severe volatility of the Bangladeshi stock market.  Macroeconomic factors such as inflation, interest rates, currency rates, and external shocks had a significant impact on it \cite{b20}.
 Investor trust has been damage and long-term economic progress has been hampered by problems including insider trading and manipulation\cite{b7}.

Because they rely on labeled data and predefined patterns, traditional regulatory and supervised machine learning approaches frequently fail to identify complex fraudulent activities in financial markets \cite{b11}. K-Means clustering has demonstrated promise in fraud detection by rapidly classifying regular and suspicious transactions \cite{b1}.
A lightweight, unsupervised, clustering-based fraud detection framework that can adaptively identify suspicious trading behaviors from evolving market data is critically needed.This study  develops a generic pipeline capable of detecting potential market manipulation and manipulative behaviors in historical stock trading data.
% , particularly when paired with methods such as Principal Component Analysis (PCA)
 
\begin{figure*}[htbp]
    \centering
    \includegraphics[width=.9\textwidth]{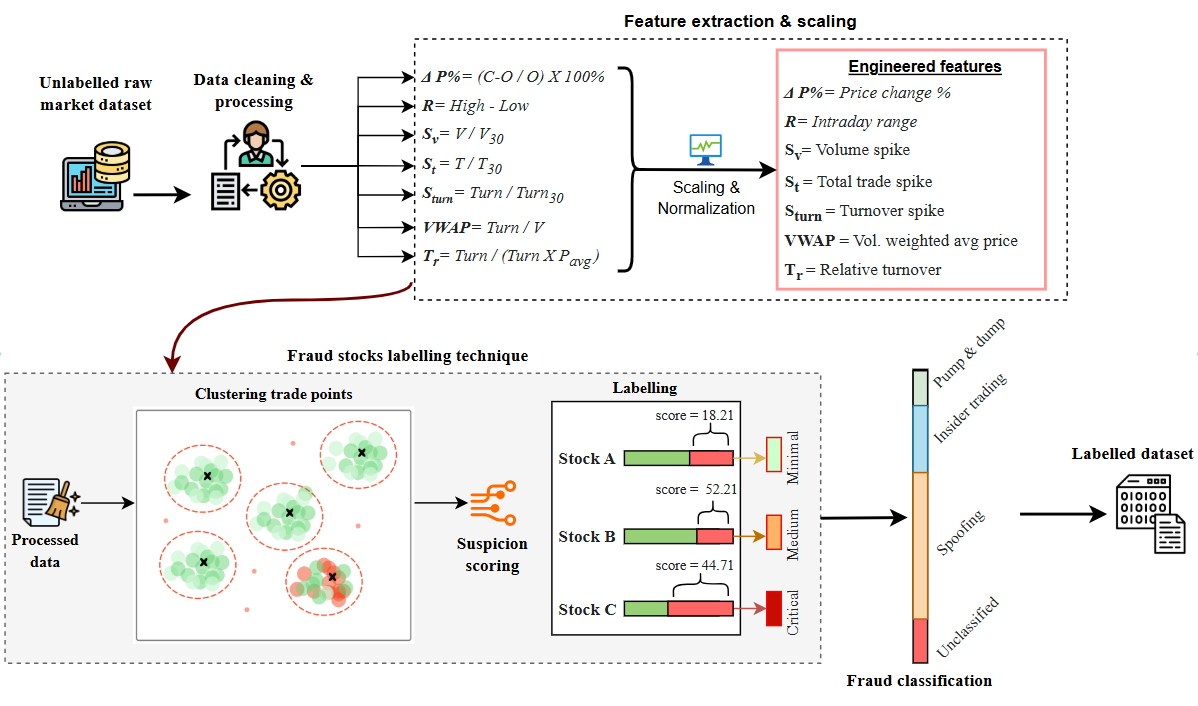}
    \caption{Market manipulation detection workflow}
    \label{fig:Market_manipulation_detection_workflow}
\end{figure*}

 The main contributions of this research include -
\begin{enumerate}

\item We propose a scalable unsupervised Stock Market Manipulation Detection (SMMD) framework that integrates clustering-based structural anomaly detection with behavioral financial heuristics.
 
\item We extract nine technical indicators using 30-day rolling windows to capture temporal patterns indicative of abnormal trading behavior.
 
\item K-Means outperforms DBSCAN, OPTICS, and hierarchical clustering in accuracy (0.987), silhouette score (0.965), and scalability, making it highly effective for detecting sparse fraudulent trading patterns \cite{b21}.The proposed KMeans clustering algorithm  marks outliers as potential indicators of suspicious activity and natural clusters of regular trading patterns.
 
\item We provide interpretable fraud-type categorization aligned with real-world manipulation patterns.
 
\item We present an adaptive percentile-based risk categorization approach and a symbol-level suspicion rating system.

\item To minimize false positives, we develop a hybrid anomaly filtering system that combines percentile-based behavioral criteria with distance-based outlier detection.
\end{enumerate}

\section{Literature Review}
Traditional supervised learning models have been widely used for fraud detection, but their effectiveness is confined by the need for enormous volumes of labeled data and inability to identify key fraud patterns. Currently, more researchers are focusing on unsupervised learning methods that could overcome these issues, especially clustering algorithms, which enable anomaly detection in unlabeled information.
\\
Zengyi et al. \cite{b1} performed actual transaction data analysis in the identification of financial fraud using the K-means unsupervised clustering technique. Principal Component Analysis was used for the simplification of data and the identification of key elements useful in fraud detection.

% SimCLR was founded by Xuan Li et al. (2024) \cite{b2} to identify anomalous activity and transaction patterns without human monitoring.  The study employed an innovative method based on SimCLR contrastive learning in addition to several data preparation strategies.  They used a shared dataset of eBay transactions that contained a variety of behaviors.

% They used SimCLR as the main technique. It's a contrastive learning method that doesn't need labeled data. We treated each transaction as a single data point and made two slightly different versions of each by adding noise or changing values. This let us mimic different viewpoints on the same transaction. Then, we fed them into a deep neural network, which made hidden data representations. The model trained using a loss function that makes similar versions of the same transaction closer together and different transactions farther apart.
SimCLR and its usage were introduced by Xuan Li et al. in 2024 \cite{b2}. SimCLR identifies transaction patterns and identifies unusual transaction activities without requiring supervision. The authors conducted an experiment on an eBay transaction data set which contained all sorts of actions. The research used various ways of data preparation and an exclusive technique which uses SimCLR contrastive learning.
% Nattaporn and Subhorn (2022) \cite{b3} attempted to solve this problem by testing the applicability of random K-means, global K-means, and fast global K-means on real-world bank data.From 594,000 transactions 7,200 were fraudulent.To identify customer clustering, three approaches were deployed: random k-means, global k-means, and fast global k-means.
%  They employed PCA to reduce the features into two dimensions.
 % To analyze how efficient it worked, they tried to identify the Sum of Squared Errors (SSE) regarding the compactness achieved by the clusters and Davies-Bouldin Index (DB Index) for separating them.

% Leangarun, Tangamchit, and Thajchayapong (2021) \cite{b8} addressed problems pertaining to stock price manipulations on the Thailand Stock Exchange which is  commonly known as SET. The authors have created two unsupervised deep learning models for identifying manipulations – LSTM Autoencoder and LSTM Generative Adversarial Networks.

To be able to detect anomalies within stock price time series data, Wenjie, Ruofan, and Bofan (2020) \cite{b10} designed a deep learning model that incorporated Conv1D and LSTM.  Conv1D enabled it to divide data into 30-day chunks. Subsequent LSTM layers enabled it to detect long-term trends. The model also incorporated three dense layers for forecasting.

% New paper 
Li et al. (2025) \cite{b3} developed an ensemble machine learning framework  to identify unusual trading by APAC investors in U.S. markets. It removed the error rates by 34.7\% and obtained an AUC-ROC of 0.971 using data from approximately 847,000 accounts across twelve APAC economies.The study emphasizes the significance of incorporating behavioral economics and advanced machine learning into RegTech.
To detect fintech fraud Darwish et al. (2025) \cite{b6} presented a multi-stage hybrid methodology. To solve class imbalance, they used rule-based filtering, K-means clustering, Artificial Bee Colony (ABC) optimization, and KNN classification with ABC-sampling.On e-commerce datasets, it obtained 95.0\% accuracy.It reduces false positives and improves robustness in fraud detection using unbalanced data.
% To address trade-and-quote market manipulation on stock markets, a solution employing machine learning methods with 22 instances of market manipulation on 20 stocks with Level 3 trade data from Borsa Istanbul (BIST) between 2010 and 2015 was adopted by Nurullah and Fuat in 2022.   Various classifiers were rendered with a maximum accuracy rate of 93\%, sensitivity rate of 95\%, and F1-score rate of 91\%, and Support Vector Machine and Naïve Bayes classifiers were seen to be most successful among these.

\section{Research Methodology}
% This study used a structured technique to detect irregular trading activity.
 An unsupervised learning model is proposed to detect the manipulated stocks of daily stock data from 2012 to 2024. Necessary steps have been taken to preprocess the data.

Fig.~\ref{fig:Market_manipulation_detection_workflow}
outlines our study's process.

\subsection{Data Preprocessing} 
% The preprocessing phase involves steps that must be conducted on the original data before the clustering model is applied. 
To ensure the quality and reliability of the stock market dataset, a series of preprocessing steps were implemented:
  \begin{itemize}
      \item The "Date" field was converted to the right datetime format, and duplicate, erroneous, and incomplete entries were eliminated from the data.
      \item Dropna() was used to eliminate  missing values in the rows, leaving a clean dataset with 1,019,783 rows.
  \end{itemize}
\subsection{Feature Engineering}
To enhance the model’s ability to detect suspicious trading patterns, multiple features were engineered from the cleaned stock dataset. 
% These features aim to capture both price behavior and trading volume dynamics, which are often early indicators of potential manipulation.
Nine new features were developed using a 30-day rolling frame to capture trading trends and detect anomalies. 
% These include the daily price change (\%), intraday range, volatility (30-day rolling average of closing price), volume spike, trade count spike, turnover spike, VWAP, relative turnover, and price-to-volume ratio.
Infinite division values were substituted with NaNs, then replaced with default values.Before being employed in anomaly detection clustering, all features were standardized.

% (1.0 for spikes/ratios, 0.0 for volatility). 

% Mainly, Feature engineering developed indicators such as Price Change, Volatility Ratio, and Volume Spike Indicator to enhance fraud detection.
\subsection{Feature Scaling}
In machine learning, feature scaling is a method for normalizing independent variables or features to a specific range\cite{b9}. After cleaning all features were consistent. Then, we standardized the selected features with StandardScaler, which gives a mean of 0 and a standard deviation of 1. This phase guarantees that all features contribute equally to the clustering procedure by increasing model performance and stability.
The pseudocode for Algorithm~\ref{alg:smmd} outlines a six-phase unsupervised pipeline to detect anomalous trading patterns in daily stock data spanning 2012--2024. The definitions of all the variables used in the pseudocode are provided in Table~\ref{tab:variables}.
\subsection{Model Training} 
% Cluster analysis, often known as 
Clustering is an unsupervised classification method that divides a set of multidimensional data into clusters based on a predetermined criterion \cite{b17}. 
% Clustering methods are often classified into two categories: probabilistic model-based approaches and nonparametric approaches\cite{b12}.
\subsubsection{K-means++ Clustering Algorithm}

K-means++ is a popular variant of K-means for minimizing the Sum of Squared Euclidean Distances (SSEDM). K-means++ iteratively selects initial cluster centres in a way that spreads them out, allowing the algorithm to converge faster and avoid poor local minima \cite{b18}.
%  It is an initialization look at for the K-means clustering algorithm that increases the final cluster quality \cite{b18}.

\subsubsection{Clustering with K-Means++ using the Elbow Method} 
To establish a k value, which is applied to cluster the datasets into k sets, we used the elbow approach to calculate the number of clusters.
% The Elbow Method helps in determining the point at which increasing the number of clusters produces decreasing rewards in terms of inertia reduction.
From the plot, we can see a significant decrease in inertia from k = 2 to 6.

\begin{figure}[tbhp]
\centerline{\includegraphics[width=0.5\textwidth]{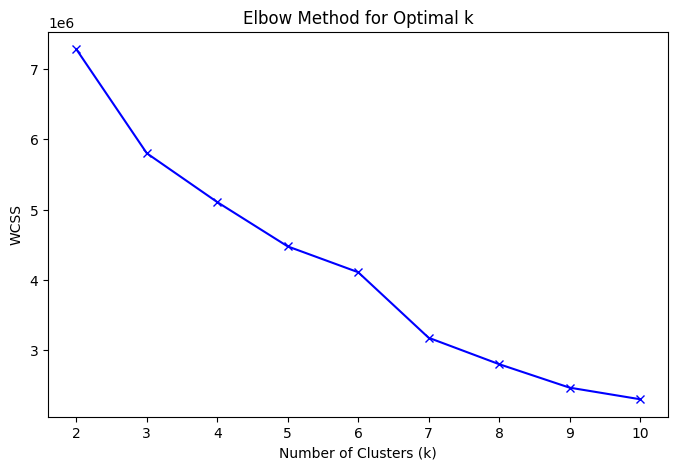}}
\caption{ Elbow Method for Optimal k}
\label{fig: Elbow_Method}
\end{figure}

%  As the inertia declines quickly until k = 6, it flattens out.
%  This "elbow" implies that k = 6 provides the best combination of compact clusters and model simplicity.
% %  , with adding more clusters providing insignificant improvement.
Fig.~\ref{fig: Elbow_Method} shows the elbow method used for optimal k clusters.

\subsubsection{K-Means++ Algorithm Implementation Steps}
This whole algorithm operates in the following steps \cite{b16}:

\begin{itemize}
    \item In order to find the value of k we used the technique of Elbow. The value of k is chosen where  the reduction in inertia slows down.
    \item The initial location of the centroid is random.   For each observation, the closest centroid in Euclidean distance is assigned. Then the center of mass is found for each of the k groups.
    \item The process is stopped when there is less than a certain fixed distance between the two centers of mass. Otherwise, the process is repeated.
\end{itemize}

\subsection{Proposed Algorithm: Stock Market Manipulation Detection (SMMD)}
\begin{algorithm}[htbp]
\caption{SMMD: Stock Market Manipulation Detection}
\label{alg:smmd}
\begin{algorithmic}[1]
\footnotesize

\Require Excel sheets 2010--2020, 2021--2024; exclude $\{2010,2011\}$; $k=5$, $p_{95}$, $\Delta p_{thr}=10$
\Ensure Flagged trades, fraud type, risk score, metrics

\Statex \textit{\textbf{// Load \& Clean}}
\State $\texttt{df} \leftarrow \Call{Concat}{\text{Load(2010--2020)}, \text{Load(2021--2024)}}$
\State Filter out years $\in \{2010,2011\}$; replace \texttt{NULL} $\to$ \texttt{NaN}
\State $\texttt{cleaned\_df} \leftarrow \Call{DropNa}{df}$

\Statex \textit{\textbf{// Features}}
\For{each row $\in$ \texttt{cleaned\_df}}
    \State $\Delta P\% \leftarrow \frac{C-O}{O}\times 100$
    \State $R \leftarrow H - L$
\EndFor

\State Group by \texttt{Symbol} $\to$ $G$
\For{each $g \in G$}
    \State $V_{30} \leftarrow \Call{RollMean}{g.\text{Vol}, 30}$
    \State $T_{30} \leftarrow \Call{RollMean}{g.\text{Trade}, 30}$
    \State $\sigma_{30} \leftarrow \Call{RollStd}{g.\text{Close}, 30}$
    \State $S_V = V/V_{30},\; S_T = T/T_{30},\; S_{Turn} = Turn/Turn_{30}$
    \State $VWAP = Turn/V,\; P_{avg} = (O+C)/2$
\EndFor

\Statex \textit{\textbf{// Scale \& Cluster}}
\State $X \leftarrow [\Delta P\%, R, \sigma_{30}, S_V, S_T, S_{Turn}, VWAP]$
\State $X \leftarrow \Call{RemoveInfNaN}{X}$
\State $X_s \leftarrow \Call{Standardize}{X}$
\State $M \leftarrow \Call{KMeans}{X_s, k}$
\State $\texttt{Cluster} \leftarrow \Call{Predict}{M}$

\Statex \textit{\textbf{// Anomaly Detection}}
\State $d_i \leftarrow \|x_i - c_{\text{cluster}}\|_2$
\State $t_d \leftarrow \Call{Quantile}{d, 0.95}$
\State $t_V, t_T \leftarrow 95^{th}$ percentile of $S_V, S_T$

\For{each row}
    \If{$d_i > t_d \land (|\Delta P\%| > 10 \lor S_V > t_V \lor S_T > t_T)$}
        \State \texttt{Suspicious} $\leftarrow$ True
    \EndIf
\EndFor

\Statex \textit{\textbf{// Risk \& Fraud Labeling}}
\State Group by \texttt{Symbol} $\to$ $G_s$

\For{each $g \in G_s$}
    \State $Score \leftarrow (\% \text{suspicious}) \times 100$
    \[
    \text{Risk} =
    \begin{cases}
    \text{Critical Risk} & Score > 90 \\
    \text{High Risk} & 75 < Score \leq 90 \\
    \text{Medium Risk} & 50 < Score \leq 75 \\
    \text{Low Risk} & 25 < Score \leq 50 \\
    \text{Minimal Risk} & Score \leq 25
    \end{cases}
    \]
    \State Sort by \texttt{Date}
    \Statex \quad $\to$ Check 5-day future: \textit{pump-and-dump}, \textit{rug pull}, \textit{spoofing}, etc.
\EndFor

\Statex \textit{\textbf{// Evaluate}}
\State $\text{Sil} \leftarrow \Call{Silhouette}{\text{Sample}(X_s \leq 10k), \text{Cluster}}$
\State Plot fraud type distribution; output ranked symbols

\end{algorithmic}
\end{algorithm}

\begin{table*}[h!tp]
\centering
\caption{Variable definitions used in the Stock Market Manipulation Detection (SMMD) pseudocode.}
\label{tab:variables}
\footnotesize
\begin{tabular}{@{}p{2.2cm}p{12.5cm}@{}} % adjusted widths for full-page
\toprule
\textbf{Symbol} & \textbf{Description} \\
\midrule
$C, O, H, L$ & Daily closing, opening, high, and low prices \\
$V, T, Turn$ & Trading volume, number of trades, and turnover (monetary value) \\
$\Delta P\%$ & Percentage price change: $\frac{C - O}{O} \times 100$ \\
$R$ & Intraday price range: $H - L$ \\
$V_{30}, T_{30}$ & 30-day rolling average of volume and number of trades \\
$\sigma_{30}$ & 30-day rolling standard deviation of closing price (volatility) \\
$S_V, S_T, S_{Turn}$ & Spike ratios: $V/V_{30}$, $T/T_{30}$, $Turn/Turn_{30}$ \\
$VWAP$ & Volume-weighted average price: $Turn / V$ \\
$P_{avg}$ & Average of open and close: $(O + C)/2$ \\
$X, X_s$ & Feature matrix and its standardized version \\
$k, M$ & Number of clusters ($k=5$) and fitted K-Means model \\
$d_i$ & Euclidean distance of point $i$ to its cluster center \\
$t_d, t_V, t_T$ & 95th percentile thresholds for distance and spike features \\
$Score$ & Suspicion score: percentage of flagged days per stock \\
\bottomrule
\end{tabular}

\label{tab:variables}
\end{table*}

\subsection{Model Evaluation}
% Silhouette Score had been applied to assess the efficiency of the system in anomaly detection. It is a measure that considers many aspects of how well the anomalous detection method is functioning.
% Silhouette Score had been applied for evaluating the effectiveness of system's anomaly detection. It assesses many aspects of the detection method's effectiveness.
Silhouette Score:
    % Silhouette analysis is used for studying the separation distance of the clusters.
Silhouette analysis mainly checks how close each item in the cluster is to items in other clusters. The score value ranges from -1 to +1. A score value of +1 indicates that objects are well-clustered, whereas a score value of -1 shows that the objects are not correctly clustered\cite{b19}.

\begin{equation}
    \begin{array}{l}
\text { For a single sample } i \text { : }\\
s(i)=\frac{b(i)-a(i)}{\max \{a(i), b(i)\}}
\end{array}
\end{equation}
Where:
\begin{enumerate}
    \item a(i) = average distance between i and all other points in the same cluster.
    \item b(i) = average distance between i and all points reside in any other cluster.
\end{enumerate}

\section{Result and Discussion}

\subsection{Anomaly Detection Results}
% The K-Means++ method form clusters with the help of Elbow Method to identify the number of clusters. 
Outliers were spotted by calculating how far each trade was from its cluster's center.The top 5\% farthest labeled as structural outliers. To detect suspicious trades, behavioral thresholds such as volume spikes, transaction count, turnover which is above the 95th percentile \cite{b14},and price fluctuations greater than 10\% were observed.
Only a trade that is distant from a cluster and exhibited at least one of the following behaviors would be considered suspicious. Lastly,the suspicious trade is processed by a fraud detection scoring system.

%  The K-Means++ method classified transactions into natural clusters based on engineered characteristics. The Elbow Method established the ideal number of clusters. Trades that were the farthest from their respective cluster centres were marked as anomalies.
% The algorithm was used to calculate the Euclidean distance between each trade and its assigned cluster centre to identify anomalies. Trades in the top 5\% of these distances were considered structural outliers. To improve identification, additional behavioural thresholds were used, such as a high volume spike, a transaction count spike, a turnover spike (all over the 95th percentile), and absolute price fluctuations of more than 10\%. A trade was considered suspicious only if it surpassed the distance threshold and met at least one of the behavioural characteristics. Each 
% discovered trade was subsequently marked as suspicious and passed to fraud detection and symbol-level scoring.
\subsection{Suspicious Symbol Scoring}
% A Suspicion Score was computed to evaluate each stock's total risk. This entailed organizing all trades by symbol and computing. The number of trades flagged as suspicious is counted as flagged trades. The number of transactions for each symbol is the total trades.
The suspicion score for stocks is calculated by combining fraud frequency and anomaly severity.The Frequency indicates if the unusual activity happen frequently and the Severity determines how extreme the unusual trades are.The formula is defined in the equation 2 :

% Requires: \usepackage{amsmath}
% \begin{equation}
%     \label{eq:placeholder_label}
%     \text{Suspicion score} = 0.6 \left( \frac{F}{T} \times 100 \right) + 0.4 \bigl( PR(D) \bigr)
% \end{equation}
\begin{equation}
    \label{eq:placeholder_label}
    \text{Suspicion score} = \alpha \left( \frac{F}{T} \times 100 \right) + (1-\alpha) \bigl( PR(D) \bigr)
\end{equation}

where:
\begin{itemize}
    \item F = Number of flagged trades.
    \item T = Total number of stock trades.
    \item PR(D) = Percentile rank of the anomaly distance.
    \item $\alpha$ = Weight parameter for frequency.
    \item $1-\alpha$ = Weight parameter for severity.
\end{itemize}

% Based on this score, each stock symbol was issued a risk label. Suspicion score, which is higher than 5\%, is in high risk, and between 3-5\% is medium risk. Less than 3\% is low risk. This rating approach helped prioritize stocks with a higher concentration of questionable trades, allowing authorities to focus on higher-risk businesses.
\subsection{Risk Categorization Method}
The next stage of the study is to classify stocks into various risk levels after calculating each stock's final suspicion score. 
%  The stocks that show the strongest signs of unusual trading behavior can be found using this classification.
  Table~\ref{tab:risk} shows the risk category and its percentile range.
\begin{table}[h!]
\centering
\caption{Adaptive Risk Categorization Based on Percentile Thresholds}
\label{tab:risk}
\begin{tabular}{|l|c|}
\hline
\textbf{Risk Category} & \textbf{Percentile Range} \\ \hline
Critical Risk & $\geq 90^{th}$  \\ \hline
High Risk     & $75^{th} - 90^{th}$  \\ \hline
Medium Risk   & $50^{th} - 75^{th}$  \\ \hline
Low Risk      & $25^{th} - 50^{th}$  \\ \hline
Minimal Risk  & $\leq 25^{th}$  \\ \hline
\end{tabular}
\end{table}
\begin{figure*}[!t]
\centering
\begin{minipage}{0.49\textwidth}
    \centering
    \includegraphics[width=\linewidth]{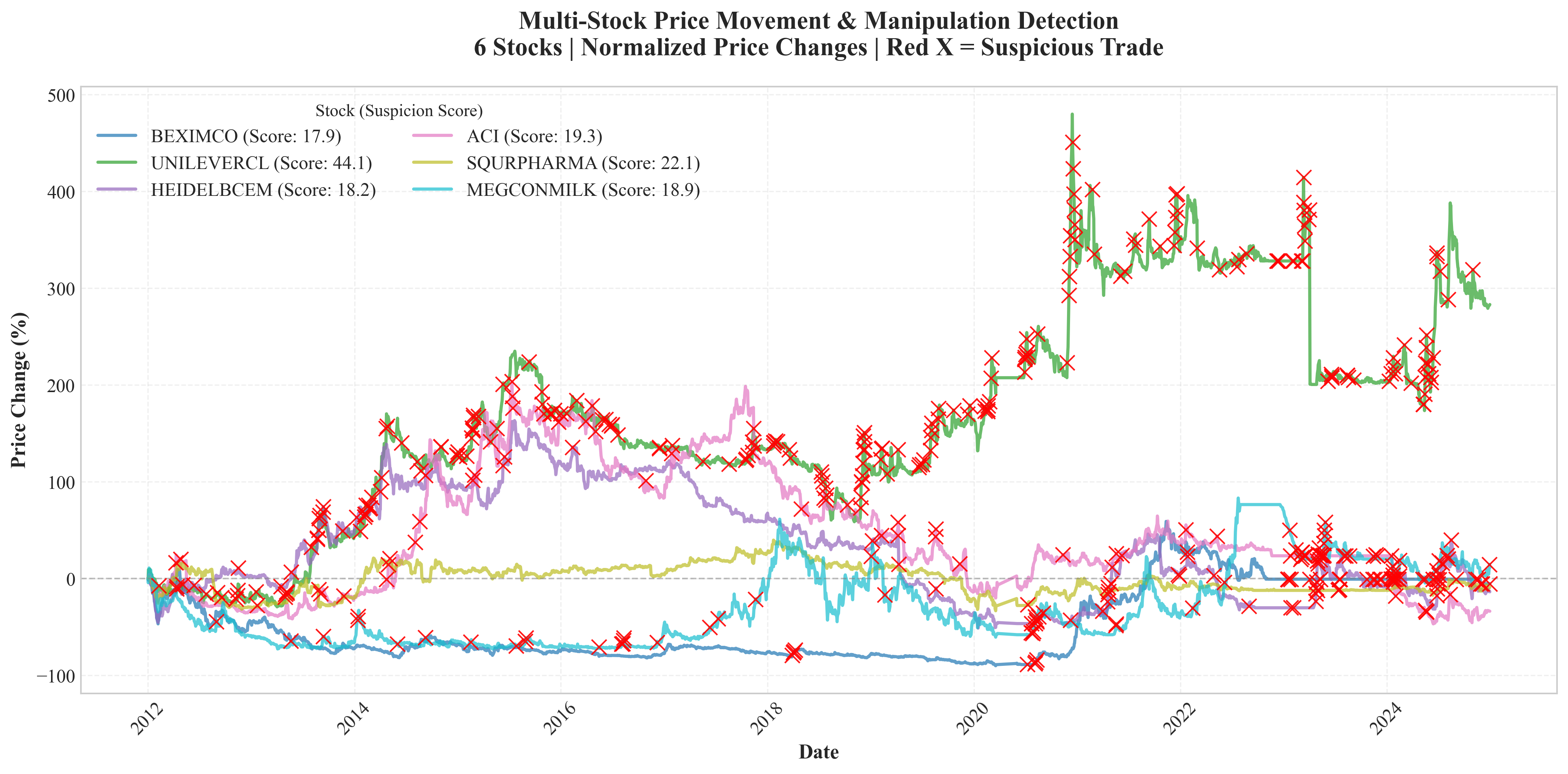}
    \caption{DSE Stocks closing price with detected suspicious trades (2012–2024)}
    \label{fig:dse}
\end{minipage}
\hfill
\begin{minipage}{0.49\textwidth}
    \centering
    \includegraphics[width=\linewidth]{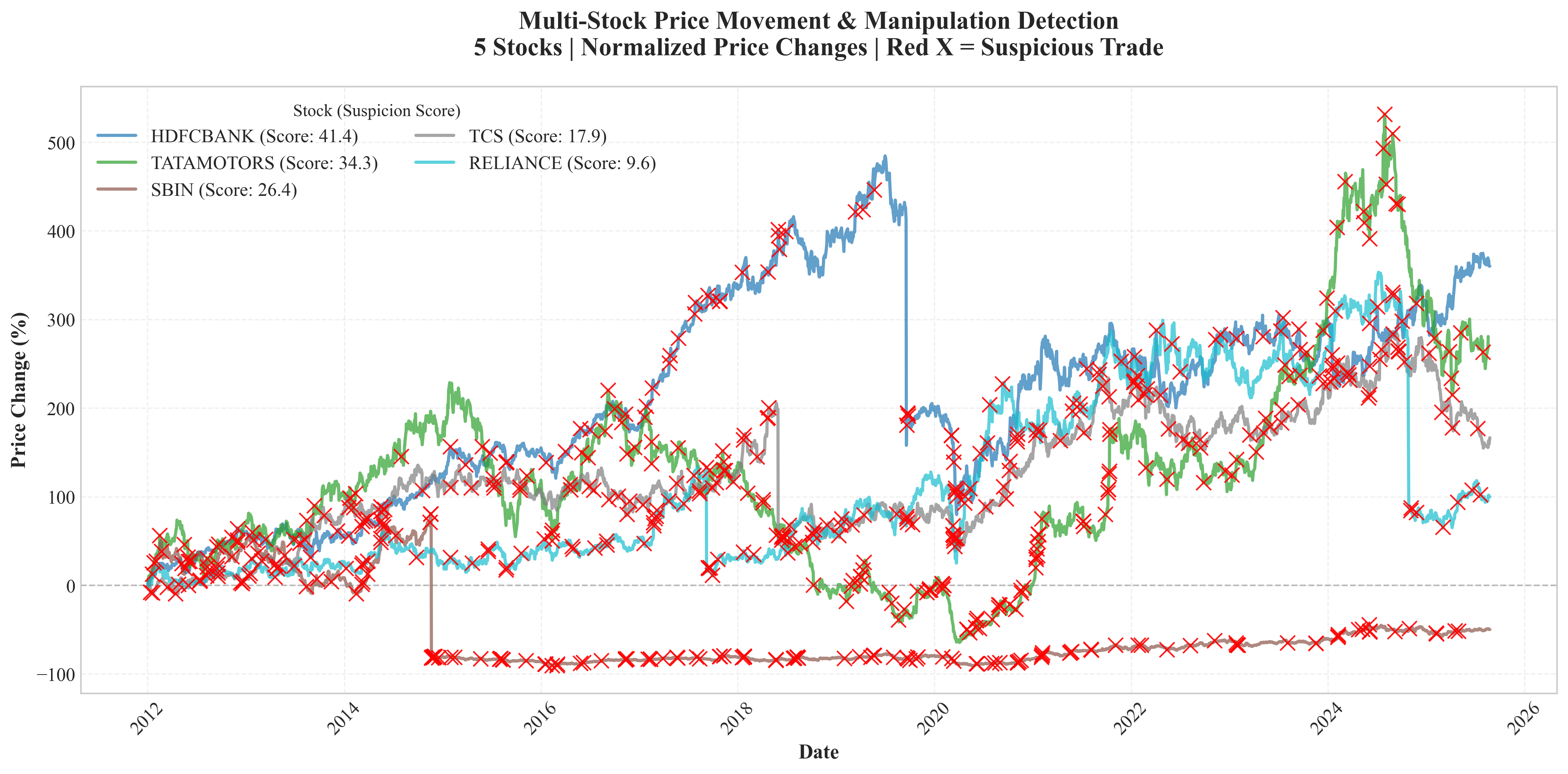}
    \caption{NSE Stocks closing price with detected suspicious trades (2012–2025)}
    \label{fig:nse}
\end{minipage}
\end{figure*}

\subsection{Fraud Type Classification} 
\begin{figure*}[htbp]
\centerline{\includegraphics[width=.9\textwidth]{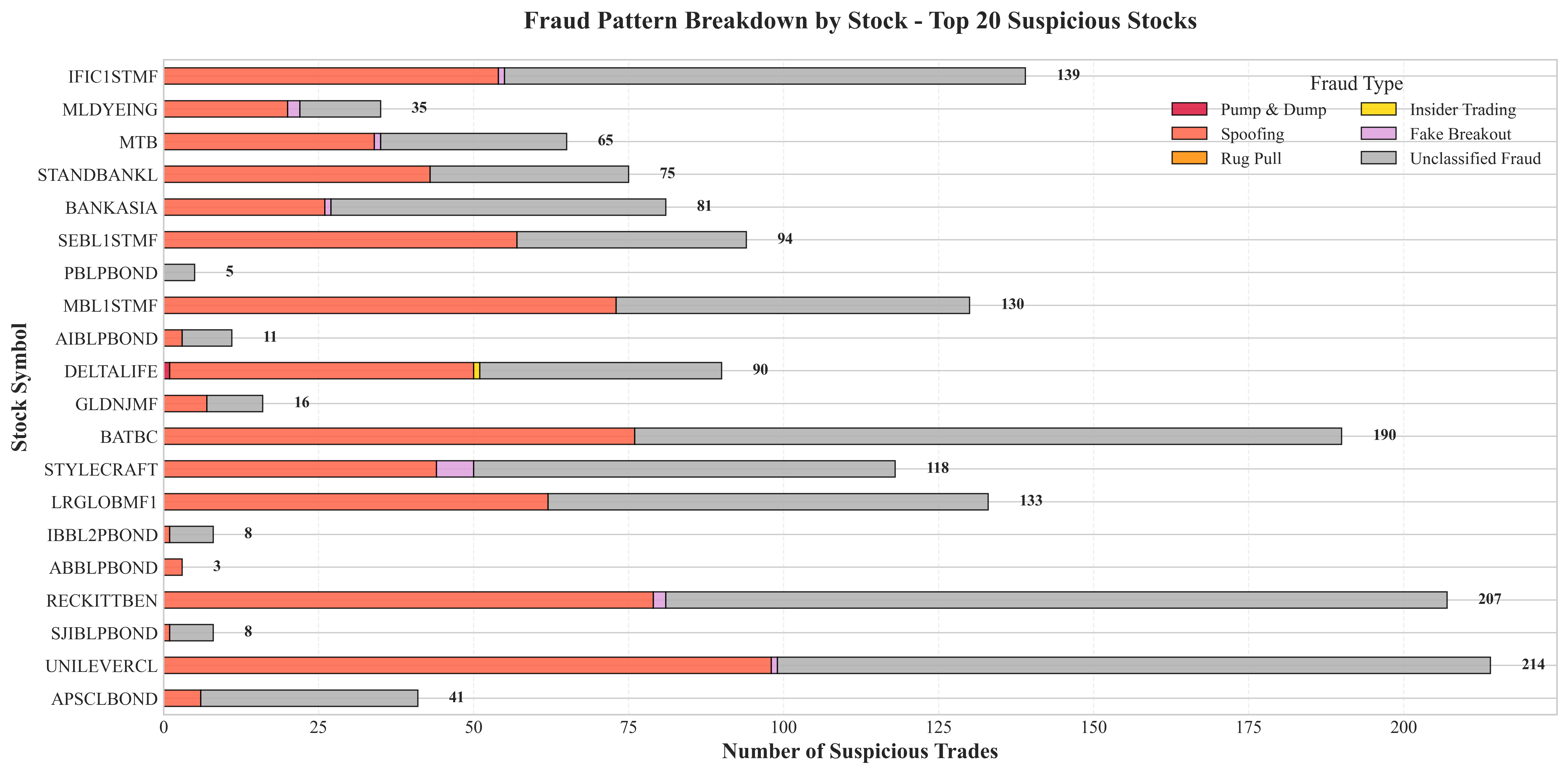}}
\caption{Fraud Pattern Breakdown of Top 20 Suspicious Stocks}
\label{fig:Fraud Pattern Breakdown of Top 20 Suspicious Stocks}
\end{figure*}

The flagged trades were examined and sorted into six fraud categories. Fig.~\ref{fig:Fraud Pattern Breakdown of Top 20 Suspicious Stocks} shows fraud pattern breakdown of top 20 suspicious stocks.

\begin{itemize}
    \item Pump \& Dump: A stock's price quickly rises by over 10\% \cite{b8} along with high trading volume, and then the price falls again within five days.

    \item Spoofing: High order or transaction volume with minor price fluctuation \cite{b12}.
    \item Rug Pull: A sharp rise in both price and volume was followed by a steep decline \cite{b13}.
    % Volume fell to less than half the moving average, and the price dropped by over ten percent within five days.

    \item Insider Trading:A sharp increase of more than ten percent on a low-volume spike of less than one point, which suggests possible insider trading.

    \item Fake Breakout: The price increased over 10\% with high volatility, but reversed within a week.

    \item Unclassified Fraud: If none of the above fits, we toss the odd data point into unclassified.  
\end{itemize}

% \begin{figure*}[htbp]
% \centerline{\includegraphics[width=.7\textwidth]{Images/marked.png}}
% \caption{DSE Stocks closing price with detected suspicious trades(2012–2024)}
% \label{fig:Stocks closing price with detected suspicious trades marked (2012–2024)}
% \end{figure*}

% \begin{figure*}[h]
% \centerline{\includegraphics[width=.7\textwidth]{Images/IndianDataset.png}}
% \caption{NSE Stocks closing price with detected suspicious trades(2012–2025)}
% \label{fig:NSE Stocks closing price with detected suspicious trades(2012–2025)}
% \end{figure*}

\subsection{Fraud Type Classification Criteria} 
Based on observable features, heuristic rules classified suspicious trading behavior to match real-world fraud patterns.
. The Table~\ref{tab:heuristic_rules} shows the particular criteria used to classify flagged trades as known fraud types.

\begin{table}[htbp]
\caption{Heuristic Rules and Feature Thresholds Used for Fraud Type Classification}
\label{tab:heuristic_rules}
\centering

% --- font size increase ---
{\fontsize{6}{6}\selectfont   % larger text (IEEE-safe)

% --- spacing in em ---
\setlength{\tabcolsep}{0 em}     % horizontal padding
\renewcommand{\arraystretch}{1.4} % vertical row spacing

\begin{tabular}{|c|c|c|c|c|c|}
\hline
\textbf{Fraud Type} & \textbf{Price (\%)} & \textbf{Vol. Spike} &
\textbf{Trades Spike} & \textbf{Range} & \textbf{Lookahead} \\
\hline
Pump \& Dump  
& $+10 \rightarrow -10$  
& $>95^{th}$ pct  
& --  
& --  
& 5 days \\
\hline
Spoofing      
& $<5$  
& $>95^{th}$ pct  
& $>95^{th}$ pct  
& --  
& Same day \\
\hline
Rug Pull      
& $+10 \rightarrow -10$  
& $>95^{th} \rightarrow <0.5\times$  
& --  
& --  
& 5 days \\
\hline
Insider Trading       
& $>10$  
& $<1.0\times$  
& --  
& --  
& Same day \\
\hline
Fake Breakout      
& $>10$ (reversal)  
& --  
& --  
& $>90^{th}$ pct  
& 5 days \\
\hline
Unclassified  
& Any  
& $>95^{th}$ pct
& $>95^{th}$ pct
& --  
& N/A \\
\hline
\end{tabular}
}
\end{table}

% From our dataset, we analyzed the percentage of different fraud types in Fig.~5.

\subsection{Visualization of fraud trades }
Each stock's closing price is plotted alongside red markers that indicate suspicious trades in the visualization.  According to their Suspicion Scores, stocks like UNILEVERCL (44.1\%), SQURPHARMA (22.1\%), ACI (19.3\%), HEIDELBCEM (18.2\%), MEGCONMILK (18.9\%), and BEXIMCO (17.9\%) display a variety of anomalous density.
% Higher scores relate with more frequently highlighted points, making it obvious where anomalous trading activity overlaps with rapid fluctuations in prices.
 To understand when these suspicious trades happened, a chart of flagged trades marked by red ‘×’ symbols.
 Fig.~\ref{fig:dse} depicts some random 6 suspicious stocks of DSE.
Our proposed manipulation detection system is applied on the NSE (National stock market of India) dataset to test its effectiveness in a global market context in  Fig.~\ref{fig:nse} .

\subsection{Model Evaluation}
 % Due to the lack of  labeled ground truth data, traditional evaluation metrics such as accuracy, precision, and recall could not be applied.
 The assessment of the clustering-based anomaly detection system relied primarily on the Silhouette Score which served as the main indicator of model performance.

The obtained scores of 0.561 and 0.292 show well-defined and separated clusters for both exchanges. 
% It also supports the appropriateness of K-Means++ to identify patterns of trading behaviors.
  
    % \item Suspicious Trade Proportion: The model identified 2.02\% of deals as suspicious. This is a reasonable rate for anomaly detection, indicating that the model is successful without causing excessive false alarms.
 
    % \item Anomaly Score Separation: On average, suspicious trades were 5.45 units distant from cluster centers, while typical deals were just 1.03 units away. This substantial discrepancy indicates that the flagged trades are clearly distinct from ordinary ones and statistically odd.

\subsection{Result Summary} 
The Clustering Model performed effectively by achieving a silhouette score of 0.561. 2.02\% of deals were reported as suspicious which indicates a suitable detection rate for identifying anomalies in DSE financial data.Spoofing was the most frequent containing 51.1\% cases while other types of fraud were less frequent.This approach should work with datasets from around the globe which remain unlabeled dataset. It worked well even in Bangladesh's unstable trading conditions.
  
% Overall, the system shows great promise as a useful fraud detection tool by successfully identifying, visualizing, and interpreting unusual trading behavior in real-world market data.
% \\
% The model could detect anomalies without labeled data. 

% The system has limits. It labels about 47\% of reported trades as Unclassified Fraud using set rules, which means it doesn't catch every kind of fraud. Also, K-Means++ uses round clusters and takes 30–80 seconds, so it's not fast enough to find fraud as it happens in high-speed trading.

% \section{Future Work}

% Future enhancements can consist of adaptive thresholding primarily based on per-symbol behavior to reduce the range of unclassified fraud instances. Advanced clustering techniques like DBSCAN or HDBSCAN can be explored to detect non-spherical fraud patterns. For real-time overall performance, algorithms like MiniBatchKMeans or frameworks like Apache Spark may be used to scale the detection device.

\section{Conclusion}
% This system spots, shows, and explains odd trading actions in actual market info and looks like it could be helpful for finding fraud. It checked over a million trades from the Dhaka Stock Exchange using rules and grouping, leading to a varied check rate of 2.02\% and a silhouette score of 0.561. It works well on volatile and irregular data, suggesting it could work on worldwide data and help label trading info that isn't structured.
The system detects, visualizes, and interprets unusual trading behavior in real-world market data. This checked more than one million trades from the Dhaka Stock Exchange with the help of rules and grouping, yielding a varying verification rate of 2.02\% with a silhouette score of 0.561. It performs well on noisy and irregular data.It works well on global data and help label trading dataset that isn't structured.
 \\The lack of confirmed ground-truth data in the study is a disadvantage.In addition,the thresholds could not adapt well to highly dynamic marketplaces. While calculating suspicion scores, the 60/40 weight distribution is a heuristic that may need to be adjusted for other trading environments.The system emphasizes classifying anomalous stock returns rather than precise classification.
Advanced clustering techniques like DBSCAN/HDBSCAN and scalable real-time processing 
% using MiniBatchKMeans or Apache Spark 
are the indications of future enhancements.The system may become more effective and flexible for market analysts and regulators by ensuring stability and market transparency in global markets.

\bibliography{references}

\end{document}